\documentclass[10pt]{iopart}

\usepackage[utf8]{inputenc} 
\DeclareUnicodeCharacter{2212}{$-$}
\usepackage[T1]{fontenc}    
\usepackage{hyperref}       
\usepackage{url}            
\usepackage{booktabs}       
\usepackage{amsfonts}       
\usepackage{nicefrac}       
\usepackage{microtype}      
\usepackage{amssymb}
\usepackage{graphicx}
\usepackage{bm}
\usepackage{psfrag}
\usepackage{subfigure}
\usepackage{color}
\usepackage[normalem]{ulem}
\usepackage{pgf}
\usepackage{pgfplots}
\usepackage{tikz}
\usetikzlibrary{decorations.pathreplacing}
\usepackage{import}
\usepackage{wrapfig}
\usepackage{pdfpages}

\newcommand{\be}{\begin{equation}}
\newcommand{\ee}{\end{equation}}
\newcommand{\ba}{\begin{eqnarray}}
\newcommand{\ea}{\end{eqnarray}}

\usepackage{appendix}

\begin{document}

\title[A jamming transition from under- to over-parametrization]{A jamming transition from under- to over-parametrization affects  generalization in deep learning\vspace{0.5em}}

\author{S Spigler$^{1,\star}$, M Geiger$^{1,\star}$, S d'Ascoli$^2$, L Sagun$^1$, G Biroli$^2$ and M Wyart$^1$}

\address{
$\phantom{.}^{1}$ Institute of Physics, \'Ecole Polytechnique F\'ed\'erale de Lausanne, 1015 Lausanne, Switzerland,\\
$\phantom{.}^{2}$ Laboratoire de Physique Statistique, \'Ecole Normale Sup\'erieure, PSL Research University, 75005 Paris, France\\\vspace{0.5em}
$\phantom{.}^{\star}$ These two authors contributed equally.
}

\vspace{10pt}
\begin{indented}
\item[]June 2019
\end{indented}

\begin{abstract}
In this paper we first recall the recent result that in deep networks a phase transition, analogous to the jamming transition of granular media, delimits the over- and under-parametrized regimes where fitting can or cannot be achieved. The analysis leading to this result support that for proper initialization and architectures, in the whole over-parametrized regime poor minima of the loss are not encountered during training,  because the number of constraints that hinders the dynamics is insufficient to allow for the emergence of stable minima. Next, we study systematically how this transition affects generalization properties of the network (i.e. its predictive power). As we increase the number of parameters of a given model, starting from an under-parametrized network, we observe for gradient descent that the generalization error displays three phases: \emph{(i)} initial decay, \emph{(ii)} increase until the transition point --- where it displays a cusp --- and \emph{(iii)} slow decay toward an asymptote as the network width diverges.  However if early stopping is used, the cusp signaling the jamming transition disappears. Thereby we identify the region where the classical phenomenon of over-fitting takes place as the vicinity of the jamming transition, and the region where the model keeps improving with increasing the number of parameters, thus organizing previous empirical observations made in modern neural networks.
\end{abstract}

\section{Introduction}

Despite the remarkable progress in designing~\cite{Lecun95,He16} and training~\cite{Ioffe15} neural networks, there is still no general theory explaining their success, and their understanding remains mostly empirical. Central questions need to be clarified, such as what conditions need to be met in order to fit data properly, why the dynamics does not get stuck in spurious local minima, and how the depth of the network affects its loss landscape.

Complex physical systems with non-convex energy landscapes featuring an exponentially large number of local minima are called glasses~\cite{reviewBB}. An analogy between deep networks and 
glasses has been proposed~\cite{dauphin2014identifying,Choromanska15}, in which the learning dynamics is expected to slow down and to get stuck in the highest minima of the loss. Yet, in the regime where the number of parameters is large  (often considered in practice), several numerical and rigorous works~\cite{Freeman16,venturi2018neural,Hoffer17,Soudry2016,Cooper18} suggest a different landscape  geometry  where the loss function is characterized by a connected level set. Furthermore, studies of the Hessian of the loss  function~\cite{Sagun16,sagun2017empirical,Ballard17} and of the learning dynamics~\cite{Lipton16,Baity18} support that the landscape is characterized by an abundance of flat directions, even near its bottom, at odds with traditional glasses. 

\subsection{Jamming transition and supervised learning}
In a previous article~\cite{Geiger18} we have introduced an analogy between supervised learning with deep neural networks and a class of glassy systems, namely random dense packings of repulsive particles. It generalized a previous seminal analogy established between the loss landscape of the perceptron (the simplest network without hidden neurons) and the energy landscape of spherical particles \cite{Franz15}, and specified the universality class to which deep learning corresponds to. The critical behavior of these granular systems, although very general, is of easier understanding when we consider particles that interact only within a finite range: upon increasing their density, such systems undergo a critical \emph{jamming transition}~\cite{Wyart05b,Liu10} when there is no longer space to accommodate all the particles  without them touching one another. Before the transition the energy is zero, and after it increases with the density. The inclusion of longer-range interactions blurs the transition but its effects are still affecting the energy landscape  \cite{Wyart05b}. Deep networks behave similarly when we look at the training loss, and, again, a clear criticality emerges when considering a ``finite-range'' loss function --- the hinge loss: when the number $P$ of training points is small enough, the network is able to learn the whole training set and reaches zero training loss, and upon increasing the dataset size we find a critical ``jamming'' point where perfect training does not occur and learning gets stuck in a positive minimum of the the training loss. 

\begin{figure}[bht]
    \centering
    \includegraphics[width=0.35\textwidth]{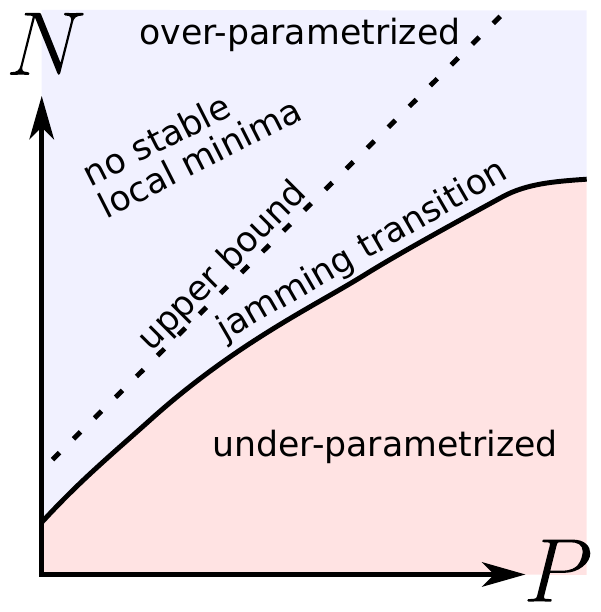}
    \caption{\small $N$: degrees of freedom, $P$: training examples. \label{fig:phasediagram}}
\end{figure}

For the full analogy we point to the aforementioned paper~\cite{Geiger18}. In the present work we first review the arguments that show that the existence of the jamming transition, studied in the $(N,P)$ plane where $N$ is the number of degrees of freedom of the network (informally, its size) and $P$ is the size of the training set. As it turns out, there is a critical line $N^\star(P)$ (whose exact location can depend on the chosen dynamics) delimiting two phases, one where the learning reaches zero training loss, and one where it gets stuck in a minimum with finite loss --- see Fig.~\ref{fig:phasediagram}. We present some numerical results that characterize the different phases, both for random data and for the MNIST dataset, using fully-connected networks with ReLU activation functions. Then, we show novel data that illustrate that this transition affects the most crucial aspect of learning, namely the generalization error. We observe that generalization properties are strongly affected by the proximity to the jamming transition: for a gradient descent dynamics, in the under-parameterized phase before jamming (large $P$ or small $N$) the generalization error is increasing; at the transition it displays a cusp; after jamming, in the over-parametrized phase, the error decreases monotonically. If early stopping is used, the cusp disappears, implying that the jamming transition is precisely the point where over-fitting is very strong.

\subsection{Generalization versus over-fitting}
The puzzle regarding the good generalization properties of neural networks despite their large size has been the topic of study for several other works. Some of them focus on the effects of various ways of regularizing the network, thereby effectively reducing the dimension \cite{caruana2001overfitting,prechelt1998early,srivastava2014dropout,krogh1992simple}. Yet another body of works focus on the effects of sheer size of a neural network \cite{neyshabur2017geometry,neyshabur2018towards,bansal2018minnorm}.

Among previous studies, \cite{advani2017high} stands out as a natural predecessor of our work. In \cite{advani2017high} the authors present one of the first empirical observations of the cusp in test error of a non-linear model, a behaviour that is reminiscent of the perceptron. There, the training dynamics is run on a two-layer student network whose training data is provided by a teacher network with a similar architecture. The authors have observed (i) a cusp in generalization error, and (ii) monotonic decay in the test error when early stopping is used.

In this work, we show  that the cusp in generalization corresponds to a phase transition  where the number of unsatisfied constraints suddenly drops to zero as $N$ increases. We quantify how the location of the transition $N^*(P)$ depends on $P$ for both random data and natural images, and find that the data structure significantly affects $N^*(P)$. Our analysis makes it clear that  $N^*\neq P$, an assumption sometimes made in previous studies. Overall, it relates the cusp in generalization to a well-known body of literature in physics associated with the ``jamming'' transition.


Since the initial preparation of the present work, the field has progressed quickly within a matter of months. The described cusp in the generalization error has been observed empirically in \cite{belkin2018reconciling} for random forest models and simple neural networks. Further theoretical studies on regression showed a precise mathematical description of the cusp behaviour in \cite{hastie2019surprises, belkin2019two,liao2018dynamics}, albeit on models that are practically somewhat further away from modern neural networks. Finally, in \cite{geiger2019scaling}, our subsequent  work, we develop a quantitative theory for (i) the cusp which is associated to the divergence of the norm of the output function at the critical point of the phase transition and (ii) the asymptotic behaviour of the generalization error as $N\rightarrow\infty$ which is associated with the reduced fluctuations of the output function. That work also shows that after ensemble averaging several networks, performance is  optimal near the jamming the threshold, emphasizing the practical importance of this transition.

\section{Theoretical framework}
\label{analogy}
In this section we recall in detail the analogy between jamming and supervised learning for deep neural networks~\cite{Geiger18,Franz15}. This will set the stage for the following thorough analysis of the phase transition and its role on generalization.

\begin{figure}[t]
    \centering
    \setlength{\unitlength}{0.1\textwidth}
    \begin{picture}(10,4.5)
    \put(-0.8,0){\scalebox{0.9}{    \begin{tikzpicture}[
        neuron/.style={line width=0.4em, white, fill=black},
        neuronX/.style={line width=0.1em, fill=white},
        synapse/.style={line width=0.01em, gray},
        synapseX/.style={line width=0.1em, black},
        scale=1.5
    ]

        \draw[decorate, decoration={brace,amplitude=10pt}] (-4.6em,-0.4em) -- (-4.6em,6.4em) node {};
        \draw (-6.3em,3.1em) node {$h$};
        \draw[decorate, decoration={brace,amplitude=10pt}] (-2.6em,1.1em) -- (-2.6em,4.9em) node {};
        \draw (-4.3em,3.1em) node {$d$};


        \draw (-1.4em,3em) node {$\mathbf{x} \rightarrow$};

        \draw[decorate, decoration={brace,amplitude=10pt}] (-0.8em,6.8em) -- (15.8em,6.8em) node {};
        \draw (7.5em,7.9em) node[above] {$L+1$};

        \draw (20.4em,3em) node[right] {$a^{(L+1)} \rightarrow f(\mathbf{x};\mathbf{W})$};

        %

        \draw[synapse] (0em,1.5em) -- (5em,0em);
        \draw[synapse] (0em,1.5em) -- (5em,1.5em);
        \draw[synapse] (0em,1.5em) -- (5em,3em);
        \draw[synapse] (0em,1.5em) -- (5em,4.5em);
        \draw[synapse] (0em,1.5em) -- (5em,6em);

        \draw[synapse] (0em,3em) -- (5em,0em);
        \draw[synapse] (0em,3em) -- (5em,1.5em);
        \draw[synapse] (0em,3em) -- (5em,3em);
        \draw[synapse] (0em,3em) -- (5em,4.5em);
        \draw[synapse] (0em,3em) -- (5em,6em);

        \draw[synapse] (0em,4.5em) -- (5em,0em);
        \draw[synapse] (0em,4.5em) -- (5em,1.5em);
        \draw[synapse] (0em,4.5em) -- (5em,3em);
        \draw[synapse] (0em,4.5em) -- (5em,4.5em);
        \draw[synapse] (0em,4.5em) -- (5em,6em);

        \draw[synapse] (5em,0em) -- (10em,0em);
        \draw[synapse] (5em,0em) -- (10em,1.5em);
        \draw[synapse] (5em,0em) -- (10em,3em);
        \draw[synapse] (5em,0em) -- (10em,4.5em);
        \draw[synapse] (5em,0em) -- (10em,6em);

        \draw[synapse] (5em,1.5em) -- (10em,0em);
        \draw[synapse] (5em,1.5em) -- (10em,1.5em);
        \draw[synapse] (5em,1.5em) -- (10em,3em);
        \draw[synapse] (5em,1.5em) -- (10em,4.5em);
        \draw[synapse] (5em,1.5em) -- (10em,6em);

        \draw[synapse] (5em,3em) -- (10em,0em);
        \draw[synapse] (5em,3em) -- (10em,1.5em);
        \draw[synapse] (5em,3em) -- (10em,3em);
        \draw[synapse] (5em,3em) -- (10em,4.5em);
        \draw[synapse] (5em,3em) -- (10em,6em);

        \draw[synapse] (5em,4.5em) -- (10em,0em);
        \draw[synapse] (5em,4.5em) -- (10em,1.5em);
        \draw[synapse] (5em,4.5em) -- (10em,3em);
        \draw[synapse] (5em,4.5em) -- (10em,4.5em);
        \draw[synapse] (5em,4.5em) -- (10em,6em);

        \draw[synapse] (5em,6em) -- (10em,0em);
        \draw[synapse] (5em,6em) -- (10em,1.5em);
        \draw[synapse] (5em,6em) -- (10em,3em);
        \draw[synapse] (5em,6em) -- (10em,4.5em);
        \draw[synapse] (5em,6em) -- (10em,6em);

        \draw[fill=white, white] (5.9em,-0.5em) rectangle (9.1em,6.5em);
        \draw (7.5em,3em) node {$\cdots$};

        \draw[synapse] (10em,0em) -- (15em,0em);
        \draw[synapse] (10em,0em) -- (15em,1.5em);
        \draw[synapse] (10em,0em) -- (15em,3em);
        \draw[synapse] (10em,0em) -- (15em,4.5em);
        \draw[synapse] (10em,0em) -- (15em,6em);

        \draw[synapse] (10em,1.5em) -- (15em,0em);
        \draw[synapse] (10em,1.5em) -- (15em,1.5em);
        \draw[synapse] (10em,1.5em) -- (15em,3em);
        \draw[synapse] (10em,1.5em) -- (15em,4.5em);
        \draw[synapse] (10em,1.5em) -- (15em,6em);

        \draw[synapse] (10em,3em) -- (15em,0em);
        \draw[synapse] (10em,3em) -- (15em,1.5em);
        \draw[synapse] (10em,3em) -- (15em,3em);
        \draw[synapse] (10em,3em) -- (15em,4.5em);
        \draw[synapse] (10em,3em) -- (15em,6em);

        \draw[synapse] (10em,4.5em) -- (15em,0em);
        \draw[synapse] (10em,4.5em) -- (15em,1.5em);
        \draw[synapse] (10em,4.5em) -- (15em,3em);
        \draw[synapse] (10em,4.5em) -- (15em,4.5em);
        \draw[synapse] (10em,4.5em) -- (15em,6em);

        \draw[synapse] (10em,6em) -- (15em,0em);
        \draw[synapse] (10em,6em) -- (15em,1.5em);
        \draw[synapse] (10em,6em) -- (15em,3em);
        \draw[synapse] (10em,6em) -- (15em,4.5em);
        \draw[synapse] (10em,6em) -- (15em,6em);

        \draw[synapse] (15em,0em) -- (20em,3em);
        \draw[synapse] (15em,1.5em) -- (20em,3em);
        \draw[synapse] (15em,3em) -- (20em,3em);
        \draw[synapse] (15em,4.5em) -- (20em,3em);
        \draw[synapse] (15em,6em) -- (20em,3em);

        \draw[neuron] (0em,1.5em) circle(0.35em);
        \draw[neuron] (0em,3em) circle(0.35em);
        \draw[neuron] (0em,4.5em) circle(0.35em);

        \draw[neuron] (5em,0em) circle(0.35em);
        \draw[neuron] (5em,1.5em) circle(0.35em);
        \draw[neuron] (5em,3em) circle(0.35em);
        \draw[neuron] (5em,4.5em) circle(0.35em);
        \draw[neuron] (5em,6em) circle(0.35em);

        \draw[neuron] (10em,0em) circle(0.35em);
        \draw[neuron] (10em,1.5em) circle(0.35em);
        \draw[neuron] (10em,3em) circle(0.35em);
        \draw[neuron] (10em,4.5em) circle(0.35em);
        \draw[neuron] (10em,6em) circle(0.35em);

        \draw[neuron] (15em,0em) circle(0.35em);
        \draw[neuron] (15em,1.5em) circle(0.35em);
        \draw[neuron] (15em,3em) circle(0.35em);
        \draw[neuron] (15em,4.5em) circle(0.35em);
        \draw[neuron] (15em,6em) circle(0.35em);

        \draw[neuron] (20em,3em) circle(0.35em);

        %

        \draw (15.7em,-1.2em) node {$a^{(L)}_\beta$};
        \draw (9.5em,-1.4em) node {$a^{(L-1)}_\alpha$};
        \draw[black] (12.5em,-1.2em) node {$W^{(L)}_{\alpha,\beta}$};
        \draw[black] (18.8em,0.3em) node {$W^{(L+1)}_\alpha$};

        \draw (5.5em,-1.2em) node {$a^{(1)}_\delta$};
        \draw (0em,0.1em) node {$x_\gamma$};
        \draw[black] (2.5em,-0.7em) node {$W^{(1)}_{\gamma,\delta}$};

    \end{tikzpicture}\vspace{0.5em}}}
    \end{picture}
    \caption{Architecture of a fully-connected network with $L$ hidden layers of constant size $h$. Points indicate neurons, connections between them are characterized by a weight. Biases are not represented here. \label{fig:architecture}}
\end{figure}
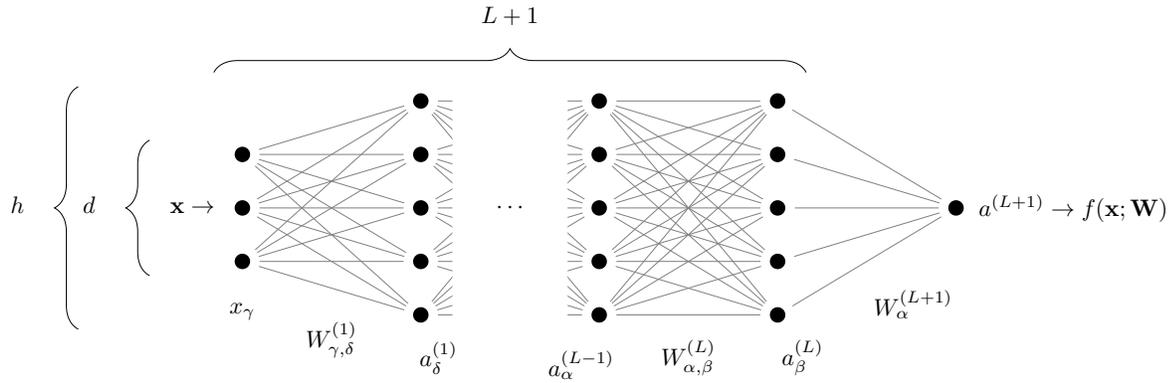

\subsection{Set-up}
We consider a binary classification problem, with a set of $P$ distinct training data denoted $\{(\mathbf{x}_\mu,y_\mu)\}_{\mu=1}^P$. The vector $\mathbf{x}_\mu$ is the input, which lives in a $d$-dimensional space, and $y_\mu=\pm 1$ is its label. We denote by $f(\mathbf{x};\mathbf{W})$ the output of a fully-connected network corresponding to an input $\mathbf{x}$, parametrized by $\mathbf{W}$. We represent the network as in Fig.~\ref{fig:architecture}, and the output function is written recursively as
\begin{eqnarray}
    f(\mathbf{x};\mathbf{W}) \equiv a^{(L+1)},\vspace{0.5em}\\
    a^{(i)}_\beta = \sum_\alpha W^{(i)}_{\alpha,\beta}\,\rho\left(a^{(i-1)}_\alpha\right) - B^{(i)}_\beta,\\
    a^{(1)}_\beta = \sum_\alpha W^{(1)}_{\alpha,\beta}\,x_\alpha - B^{(1)}_\beta,
    \label{eq:recnnet}
\end{eqnarray}
where $a^{(i)}_\alpha$ are the preactivations. In our notation the set of parameters $\mathbf{W}$ includes, with a slight abuse of notation, both the weights $W^{(i)}_{\alpha,\beta}$ and the biases $B^{(i)}_\alpha$. $\rho(z)$ is the non-linear activation function, e.g. the ReLU $\rho(z) = z \theta(z)$ or the hyperbolic tangent $\rho(z)=\tanh(z)$. The parameters are learned by minimizing the quadratic hinge loss:
\vspace{-0.35em}
\begin{equation}
    \mathcal{L}(\mathbf{W}) = \frac1P \sum_{\mu=1}^P \frac12 \mathrm{max}\left(0, \Delta_\mu\right)^2 \equiv \frac1P \sum_{\mu\in m} \frac12 \Delta_\mu^2,
\end{equation}
where $\Delta_\mu \equiv 1 - y_\mu f(\mathbf{x}_\mu;\mathbf{W})$ and $m$ is the set of patterns with $\Delta_\mu>0$ and contains $N_\Delta$ elements. These patterns describe \emph{unsatisfied constraints}: they are either incorrectly classified or classified with an insufficient margin (whereas patterns with $\Delta_\mu<0$ are learned with margin 1). We adopt this loss function since it makes the jamming transition simpler to analyze\footnote{The often used cross-entropy loss function also displays a transition where all data are well-fitted. However, in the over-parametrized regime the dynamic never stops, as the total loss  vanishes only if the output and therefore the  weights diverge. Imposing a time cut-off is done in practice, but it blurs the criticality near jamming, as exemplified below with the early stopping procedure.}, but this choice does not influence the performance of the network, as we have reported in~\cite{Geiger18}.

We are interested in the transition between an over-parametrized phase where the network can satisfy all the constraints ($\mathcal{L}=0$) and an under-parametrized phase where some constraints remain unsatisfied ($\mathcal{L}>0$).

\subsection{A note on the effective number of parameters}
In our discussion the  notion of effective number of degrees of freedom is important.  In the space of functions going from the neighborhoods of the training set to real numbers, consider the manifold of functions $f(\mathbf{x};\mathbf{W})$ obtained by varying $\mathbf{W}$.  We denote by $N_\mathrm{eff}(\mathbf{W})$ the dimension of the tangent space of this manifold at $\mathbf{W}$. In general we have $N_\mathrm{eff}(\mathbf{W})\leq N$. Several reasons can make $N_\mathrm{eff}(\mathbf{W})$ strictly smaller than  $N$, including:

\begin{itemize}

\item The signal does not propagate in the network, i.e. $f(\mathbf{x};\mathbf{W})=C_1$ for all $\mathbf{x}$ in the neighborhood of the training points $\mathbf{x}_\mu$.
In that case, the manifold is of dimension unity and $N_\mathrm{eff}(\mathbf{W})=1$. 
This situation will occur for a poor initialization of the weights, for example if all biases are too negative on the neurons of one layer  for ReLU activation function (see for instance~\cite{schoenholz2016deep}). It can also occur if the data $\mathbf{x}_\mu$ are chosen in an adversarial manner  for a given choice of initial weights. For example, one can choose input patterns so as to not activate the first layer of neurons (which is possible if the number of such neurons is not too large). Poor transmission will be enhanced (and adversarial choices of data will be made simpler) if the architecture presents some bottlenecks. In the situation where $N_\mathrm{eff}(\mathbf{W})=1$, it is very simple to obtain local minima of the loss at finite loss values, even when the model has many parameters.

\item The activation function is linear. Then, the output is an affine function of the input, leading to $N_\mathrm{eff}\leq d+1$. Dimension-dependent bounds will also exist if the activation function is polynomial (because the output function then is also restricted to be polynomial).

\item
Symmetries are present in the network, e.g. the scale symmetry in ReLU networks: since the ReLU function is homogeneous, multiplying the weights of a layer by some factor and dividing the weights in the next layer by the same factor leaves the output function invariant. It will reduce one degrees of freedom per node. 

\item
Some neurons are never active e.g. in the ReLU case, their associated weights do not contribute to $N_\mathrm{eff}$.

\end{itemize}

Thus there are $N-N_\mathrm{eff}$ directions in parameter space that do not affect the function. These directions will lead to zero modes in the Hessian at any minimum of the loss. In what follows we consider stability with respect to the relevant $N_\mathrm{eff}$ directions, which do affect the output function. Our results on the impossibility to get stuck in bad minima are  expressed in terms of $N_\mathrm{eff}$. However, as reported in~\ref{app:neff}, we find empirically that for a proper initialization of the weights and constant-width fully connected networks, $N_\mathrm{eff}\approx N$ (the difference is small and equal to the number of hidden neurons, and only results from the symmetry associated with each ReLU neuron). Henceforth to simplify notations we will use the symbol $N$ to represent the number of effective parameters.

\subsection{Constraints on the stability of minima}
\label{sec:stability}
In this section we show that the existence of a minimum at a vanishingly small training loss (i.e. approaching jamming) is enough to derive an upper bound for the transition in the $(N,P)$ plane.

Let us suppose (and justify later) that, for a fixed number of data $P$ and with proper initialization of weights, if $N$ is large enough then gradient descent leads to ${\cal L}=0$, whereas if $N$ is small after training ${\cal L}>0$. Imagine increasing $N$ starting from a small value: at some $N^*$ the loss obtained after training approaches zero \footnote{For finite $P$, $N^*$ will present fluctuations induced by differences of initial conditions. The fluctuations of $P/N^*$  are however expected to vanish in the limit where $P$ and $N^*$ become large. This phenomenon is well-known for the jamming of particles, and is an instance of finite size effects. }, i.e. $\lim_{N\rightarrow N^*}{\cal L}=0$. We refer to this point as the jamming transition. A vanishing training loss implies that $\Delta_\mu  \rightarrow 0$ for each pattern $\mu=1,\dots, P$.
As argued in \cite{Tkachenko99}, for each $\mu \in m$ the constraint $\Delta_\mu \approx 0$ defines a manifold of dimension $N-1$\footnote{Related  arguments were recently made for a quadratic loss \cite{Cooper18}. In that case, we expect the landscape to be related to that of floppy spring networks, whose spectra are predicted in \cite{During13}.}. Satisfying $N_\Delta$ such equations thus generically leads to a manifold of solutions of dimension $N-N_\Delta$\footnote{Note that this argument implicitly assumes that  the $N_\Delta$ constraints are independent. In disordered systems this assumption is generally correct, but it may break down if  symmetries are present.}. Imposing that a solution exists implies that at jamming:
\be
N^* \geq N_\Delta.
\ee

{\it Smooth activation function:} An opposite bound can  be obtained by considerations of stability (as was done for the jamming of repulsive spheres in \cite{Wyart05a}), by imposing that in a stable minimum the Hessian must be positive definite if the output function is smooth, as it must be the case if the activation function is smooth (see below for the situation where the function displays cusps, as occurs for ReLU neurons). 
The Hessian matrix, that is the matrix of second derivatives, is
\begin{eqnarray}
\label{3}
{\cal H}_{\cal L} &= \frac1P \sum_{\mu \in m} \nabla\Delta_\mu \otimes \nabla\Delta_\mu + \frac1P \sum_{\mu \in m} \Delta_\mu \nabla\otimes\nabla \Delta_\mu \nonumber \\
&\equiv {\cal H}_0 + {\cal H}_p.
\end{eqnarray}
(Here $\nabla$ is the gradient operator and $\otimes$ stands for tensor product). The first term ${\cal H}_0$ is positive semi-definite: it is the sum of $N_\Delta$ rank-one matrices, thus ${\rm rk}({\cal H}_0) \leq N_\Delta$, implying that the kernel of ${\cal H}_0$ is at least of dimension $N-N_\Delta$. 

Let us denote by $E_-$ the negative eigenspace\footnote{The negative eigenspace is the subspace spanned by the eigenvectors associated with negative eigenvalues.} of ${\cal H}_p$ and call $N_-$ its dimension. 
Stability then imposes that $\mathrm{ker}({\cal H}_0) \cap E_- = \{0\}$, which is only possible if $N_\Delta \geq N_-$. Hence, minima with positive training loss (and therefore the minimum found right above jamming) can only occur for:

\begin{equation}
    \label{4bis}
    P\geq N_{\Delta} \geq N_-.
\end{equation}
(The first inequality trivially follows from the fact that the $N_\Delta$ patterns belong to the training set of size $P$). 
As reported in \cite{Geiger18}, we observe empirically that the spectrum of $\mathcal{H}_p$ is statistically symmetric in the cases that we consider in the present work, i.e. for ReLU activation function, both for MNIST and random data, both at initialization and at the end of training. In~\ref{app:hpsym} we provide a non-rigorous argument supporting that in the case of ReLU activation functions and random data the spectrum of $\mathcal{H}_p$ is indeed symmetric with $\lim_{N\rightarrow\infty} N_-/N_+=1$  independently of depth, where $N_+$ is the number of positive eigenvalue.
We conjecture that in general the limiting spectrum of $H_p$ as $N, P \rightarrow \infty$ (for any fixed ratio $P/N$) has a finite fraction $C_0 = N_- / N$ of negative eigenvalues for generic architectures and datasets.
In ~\cite{Geiger18} we observed $C_0 = 1/2$ for the ReLU activation function as expected, for tanh activation functions at jamming and at the end of training we found $C_0 \approx 0.43$. Thus,  $C_0$ is not universal.

Finally we assume that the spectrum of $H_p$ does not display a finite density of zero eigenvalues (once restricted to the space of parameters that affect the output, of dimension $N_\mathrm{eff}$, supposed here to be equal to $N$). Note that this assumption breaks down if data can be identical with different labels, a case we exclude here \footnote{ Indeed even in the over-parametrized case, if $\mathbf{x}_i=\mathbf{x}_j$ but $y_i\neq y_j$ then an exact cancellation of terms occurs in the sum defining $H_p$, which can then be zero while ${\cal L}>0$.}. Under this assumption we obtain from Equation~(\ref{4bis}) that local minima with positive training loss cannot be encountered if $N>P/C_0$, implying in particular that:
\be
N^*\leq  P/C_0
\ee

{\it Non-smooth activation functions:} With ReLU activation functions, the output function $f(\mathbf{x}; \mathbf{W})$ is not smooth and presents cusps, so that the Hessian needs not be positive definite for stability. A minimum can lie on a point where the second derivative is not defined along some directions (because of the cusp), and we say that the cusp stabilizes those directions. Equation~(\ref{4bis}) needs to be modified accordingly: introducing the number of directions $N_c\equiv \beta N$ presenting cusps near jamming, stability implies  $N_\Delta>N_--N_c$ and:
\be
N_\Delta\geq N(C_0-\beta)
\ee
implying in turn that:
\be
N^*(P)\leq \frac{P}{C_0-\beta}
\ee
 Numerically, we find that at jamming the fraction of directions along which there is a cusp is $N_c/N^*\equiv \beta \in (0.21, 0.25)$ both for random data and images as reported in the \ref{app:zeros}. Using $C_0=1/2$ for Relu, we obtain the bounds:
\ba
\label{111}
N_\Delta&\geq& N/4 \\
N^* &\leq& 4P.
\ea

{\it Main results:} Overall,  our analysis supports that for smooth activation functions there exists a constant $C_0$ such that:
\begin{itemize}
    \item there is a  transition for $N^*(P)\leq  P/C_0$ below which  the training loss converges to some non-zero value (under-parametrized phase) and above which  it becomes null (over-parametrized phase).
    \item At the transition, the fraction $N_\Delta/N$ of unsatisfied constraints per degree of freedom jumps discontinuously to a finite value satisfying  $C_0\leq N_\Delta/N\leq 1$.
\end{itemize}
The complete list of results, including consequences of this analysis on the Hessian, is included in \cite{Geiger18}.

For $Relu$ activation function,  $C_0=1/2$ but the analysis is complicated by the presence of cusps. The jamming transition is still sharp, i.e.~characterized by a discontinuous jump in constraints as specified by  Eq.\ref{111}.

In the  next sections, we confirm these predictions for $Relu$ in numerical experiments and observe the generalization properties at and beyond the transition point.





\begin{figure*}[t!]
    \centering
    \setlength{\unitlength}{0.1\textwidth}
    \begin{picture}(10,4)
    \put(-1.5,0){\scalebox{0.68}{\import{figures/}{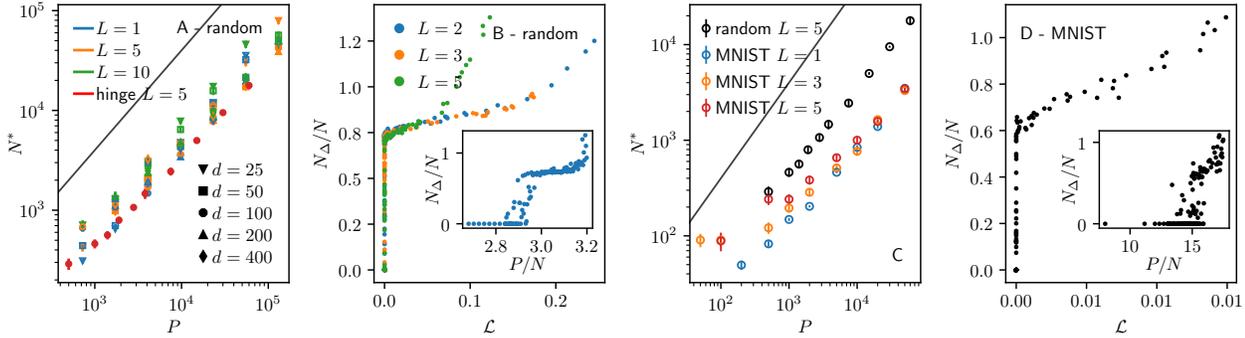}}}
    \end{picture}
    \caption{\small (A, B) Random data and (C, D) MNIST dataset. (A) and (C) depict the location $N^*$ of the transition as a function of the number $P$, for networks with different cost functions or sizes. (B) and (D) show that in the $\mathcal{L}$-$N_\Delta/N$ and $P/N$-$N_\Delta/N$ planes the transition displays a discontinuous jump.}
    \label{fig:fit}
\end{figure*}

\section{Location of the jamming transition}
\label{sec:random}
Here we present the numerical results on random data (uniformly distributed on a hypersphere and with random labels $y_\mu=\pm 1$) and on the MNIST dataset (partitioned into two groups according to the parity of the digits and with labels $y_\mu=\pm1$). With MNIST, in order not to have most of the weights in the first layer, we reduce the actual input size by retaining only the first $d=10$ principal components that carry the most variance (this hardly diminishes the performance for such a task). Further description of the protocols is in \ref{app:simulations}.

In Fig.~\ref{fig:fit}A,C we show the location of boundary $N^*$ versus the number of samples $P$. $N^*$ is estimated numerically for each $P$ by starting from a large value of $N$ and progressively decreasing it until $L>0$ at the end of training.
Varying input dimension, depth and loss function (cross entropy or hinge) has little effect on the transition. This result indicates that in the present setup the ability of fully-connected networks to fit random data does not depend 
crucially on depth. Fig.~\ref{fig:fit}C shows also a comparison of random data with MNIST. A difference between random data and images is that the minimum number of parameters $N^*$ needed to fit the real data is significantly smaller and grows less fast as $P$ increases --- for $P\gg1$, $N^*(P)$ could be sub-linear or even tend to a finite asymptote: how the data structure affects $N^*(P)$ is an important questions for future studies.

From the analysis of Section~\ref{analogy}, the number of constraints per parameter $N_\Delta/N$ is expected to jump discontinuously at the transition. This is shown in the insets of Fig.~\ref{fig:fit}B,D. The scatter in these plots presumably reflects finite size effects known to occur near the jamming transition of particles \cite{Liu10}. All this scatter is however gone when plotting $N_\Delta/N$ as a function of the loss itself, as shown in the main panels of Fig.~\ref{fig:fit}B,D.

\begin{figure*}[b]
    \centering
    \setlength{\unitlength}{0.1\textwidth}
    \begin{picture}(10,4)
    \put(-1.5,0){\scalebox{0.68}{\import{figures/}{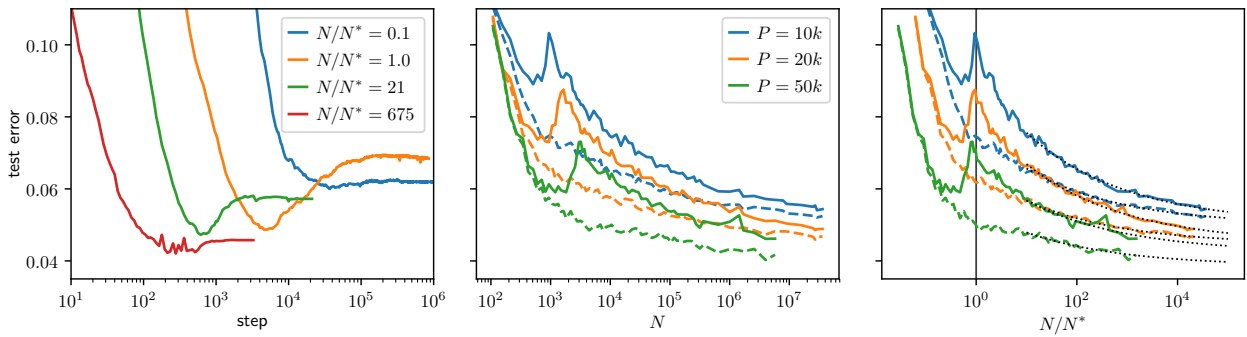}}}
    \end{picture}
    \caption{\small We trained a 5 hidden layer fully-connected network on MNIST.  (A) Typical evolution of the generalization error over training time, for systems located at different points relatively to the jamming transition (for $P=50k$): over-fitting is marked by the gap between the value at the end of training and the minimum at prior times. Notice that training of over-parametrized systems halts sooner because the networks have achieved zero loss over the training set. (B) Test error at the final point of training (solid line) and minimum error achieved during training (dashed line) vs. system size. (C) When $N$ is scaled by $N^*(P)$ it is clear that over-fitting occurs at the jamming transition. 
    }.
    \label{fig:gen}
\end{figure*}

\section{Generalization at and beyond jamming}
\label{sec:gen}

In Fig.~\ref{fig:gen}A we show the evolution of the generalization error for networks at four different locations in the $(N,P)$ plane. The networks are trained on MNIST at fixed $P=50k$, and at different values $N$, both above, at and below jamming. Training is run for a fixed number of steps of vanilla  gradient descent (the simulation details are in \ref{app:simulations}). The profile of these curves is typical of most learning problems (if one does not recur to early stopping): notice that the point of minimum generalization error happens before the end of training. The increase of test error at late times is referred  to ``over-fitting'' in the field.  Very interestingly, it is clear from this figure that at small and large $N$, over-fitting is a weak effect, which however becomes very significant at intermediate $N$.

To study this effect, we systematically vary $N$ at fixed $P$. In Fig.~\ref{fig:gen}B the solid curve shows the generalization error against the network size $N$ for three different values of $P$ (we sampled subsets of MNIST). The dashed curve represents the value of the smallest error obtained during training, at prior time-steps (extracted from the profiles shown in Fig.~\ref{fig:gen}A). The former displays a cusp at the transition point, as one can see clearly after rescaling the $N$-axis of each curve by the corresponding value of $N^\star(P)$.
Strong over-fitting, corresponding to the difference between the solid and dashed lines,  takes place only in the vicinity of the critical jamming transition (Fig.~\ref{fig:gen}B-C). We thus  posit that at fixed $P$, the benefit of early stopping~\cite{prechelt1998early} should diminish in the large-size limit. Beyond the jamming point, the accuracy keeps steadily improving as the number of parameters increases~\cite{neyshabur2017geometry,neyshabur2018towards,bansal2018minnorm}, although it does so quite slowly. We have provided a quantitative explanation for this phenomenon in \cite{geiger2019scaling}. In \ref{app:realdatadepths} we have verified that the overall trends showed in Fig.~\ref{fig:gen} qualitatively hold also for other depths.

Notice that although the cusp has been found also in shallow networks (in particular the perceptron \cite{saad1995line,engel2001statistical}), their behavior is at odds with what we observe: for the perceptron, test error asymptotically \emph{increases} with $N$.




\section{Conclusions}
Understanding the effect of over-parametrization on the behavior of deep neural networks is a central problem in machine learning.   
In this work, by focusing on the hinge loss, we recast the minimization of the loss function as a constraint-satisfaction problem with continuous degrees of freedom. A similar approach was used in the field of interacting particles, which display a sharp jamming transition affecting the landscape if the  interaction is chosen to be finite range \cite{Liu10}. Following the analogy we were able to predict a sharp transition as the number of network parameters is varied, separating a region in the $(P,N)$ plane where a global minimum can be found ($\mathcal{L}=0$) from a region where the number of unsatisfied constraints is a fraction of the number of parameters, so $\mathcal{L}>0$. These results also shed light on several aspects of deep learning:

{\it Not getting stuck in local minima:} In the over-parametrized regime, the dynamics does not get stuck in local minima at finite loss value because the number of constraints to satisfy is too small to hamper minimization. It follows from our assumptions on the negative eigenspace of the matrix $\mathcal{H}_p$ that in this regime the landscape is flat and local minima do not exist (assuming that the number of effective parameters that affect the output function is $N$).  For a smooth activation function  we predict that one cannot get stuck in a bad minimum for $N\geq P/C_0$, implying in particular that $N^*\leq P/C_0$ where $C_0$ is a constant.  We obtain a less demanding bound for ReLU activation functions due to the presence of cusps in the landscape, a situation for which we expect $C_0=1/2$. In practice, for random data $N^*(P)$ scale linearly with $P$ (in this sense, the bound is tight). By contrast, for structured data $N^*(P)$ appears to scale sub-linearly  with $P$. Predicting the curve $N^*(P)$ remains a challenge for the future.

{\it Reference point for fitting and generalization:} There exists a critical curve $N^*(P)$ on the $N$-$P$ plane above which the global minima of the landscape become accessible. The curve also appears to be linked to the generalization potential of the model. We show that in the cases that we considered, \emph{(i)} the generalization error decreases when $N\ll N^*$; then \emph{(ii)} it increases and culminates in a cusp at $N\approx N^*$ that is erased by early stopping, most useful in this region; finally, \emph{(iii)} in the over-parametrized phase, it monotonically decreases, although very slowly.


\subsubsection*{Acknowledgments}

We thank Marco Baity-Jesi, Carolina Brito, Chiara Cammarota, Taco S. Cohen, Silvio Franz, Yann LeCun, Florent Krzakala, Riccardo Ravasio, Andrew Saxe, Pierfrancesco Urbani and Lenka Zdeborova for helpful discussions. This work was partially supported by the grant from the Simons Foundation (\#454935 Giulio Biroli, \#454953 Matthieu Wyart). M.W. thanks the Swiss National Science Foundation for support under Grant No. 200021-165509. The manuscript ~\cite{franzpre}, which appeared at the same time as ours, shows that the critical properties of the jamming transition found for the non-convex perceptron \cite{Franz16} hold more generally in some shallow networks. This universality is an intriguing result. Understanding the connection with our findings
is certainly worth future studies.


\vspace{4em}

\bibliography{main}{}

\begin{thebibliography}{10}

\bibitem{Lecun95}
Yann LeCun, Yoshua Bengio, et~al.
\newblock Convolutional networks for images, speech, and time series.
\newblock {\em The handbook of brain theory and neural networks},
  3361(10):1995, 1995.

\bibitem{He16}
Kaiming He, Xiangyu Zhang, Shaoqing Ren, and Jian Sun.
\newblock Deep residual learning for image recognition.
\newblock In {\em Proceedings of the IEEE conference on computer vision and
  pattern recognition}, pages 770--778, 2016.

\bibitem{Ioffe15}
Sergey Ioffe and Christian Szegedy.
\newblock Batch normalization: Accelerating deep network training by reducing
  internal covariate shift.
\newblock In {\em International conference on machine learning}, pages
  448--456, 2015.

\bibitem{reviewBB}
Ludovic Berthier and Giulio Biroli.
\newblock Theoretical perspective on the glass transition and amorphous
  materials.
\newblock {\em Reviews of Modern Physics}, 83(2):587, 2011.

\bibitem{dauphin2014identifying}
Yann~N Dauphin, Razvan Pascanu, Caglar Gulcehre, Kyunghyun Cho, Surya Ganguli,
  and Yoshua Bengio.
\newblock Identifying and attacking the saddle point problem in
  high-dimensional non-convex optimization.
\newblock In {\em Advances in Neural Information Processing Systems}, pages
  2933--2941, 2014.

\bibitem{Choromanska15}
Anna Choromanska, Mikael Henaff, Michael Mathieu, G{\'e}rard Ben~Arous, and
  Yann LeCun.
\newblock The loss surfaces of multilayer networks.
\newblock In {\em Artificial Intelligence and Statistics}, pages 192--204,
  2015.

\bibitem{Freeman16}
C~Daniel Freeman and Joan Bruna.
\newblock Topology and geometry of deep rectified network optimization
  landscapes.
\newblock {\em International Conference on Learning Representations}, 2017.

\bibitem{venturi2018neural}
Luca Venturi, Afonso Bandeira, and Joan Bruna.
\newblock Neural networks with finite intrinsic dimension have no spurious
  valleys.
\newblock {\em arXiv preprint arXiv:1802.06384}, 2018.

\bibitem{Hoffer17}
Elad Hoffer, Itay Hubara, and Daniel Soudry.
\newblock Train longer, generalize better: closing the generalization gap in
  large batch training of neural networks.
\newblock In {\em Advances in Neural Information Processing Systems}, pages
  1729--1739, 2017.

\bibitem{Soudry2016}
Daniel Soudry and Yair Carmon.
\newblock No bad local minima: Data independent training error guarantees for
  multilayer neural networks.
\newblock {\em arXiv preprint arXiv:1605.08361}, 2016.

\bibitem{Cooper18}
Yaim Cooper.
\newblock The loss landscape of overparameterized neural networks.
\newblock {\em arXiv preprint arXiv:1804.10200}, 2018.

\bibitem{Sagun16}
Levent Sagun, {L{\'e}on} Bottou, and Yann LeCun.
\newblock Singularity of the hessian in deep learning.
\newblock {\em International Conference on Learning Representations}, 2017.

\bibitem{sagun2017empirical}
Levent Sagun, Utku Evci, V.~{U\u{g}ur} {G\"uney}, Yann Dauphin, and {L\'eon}
  Bottou.
\newblock Empirical analysis of the hessian of over-parametrized neural
  networks.
\newblock {\em ICLR 2018 Workshop Contribution, arXiv:1706.04454}, 2017.

\bibitem{Ballard17}
Andrew~J Ballard, Ritankar Das, Stefano Martiniani, Dhagash Mehta, Levent
  Sagun, Jacob~D Stevenson, and David~J Wales.
\newblock Energy landscapes for machine learning.
\newblock {\em Physical Chemistry Chemical Physics}, 2017.

\bibitem{Lipton16}
Zachary~C Lipton.
\newblock Stuck in a what? adventures in weight space.
\newblock {\em International Conference on Learning Representations}, 2016.

\bibitem{Baity18}
Marco Baity-Jesi, Levent Sagun, Mario Geiger, Stefano Spigler, Gerard~Ben
  Arous, Chiara Cammarota, Yann LeCun, Matthieu Wyart, and Giulio Biroli.
\newblock Comparing dynamics: Deep neural networks versus glassy systems.
\newblock In {\em Proceedings of the 35th International Conference on Machine
  Learning}, pages 314--323, 2018.

\bibitem{Geiger18}
Mario Geiger, Stefano Spigler, {St{\'e}phane} d'Ascoli, Levent Sagun, Marco
  Baity-Jesi, Giulio Biroli, and Matthieu Wyart.
\newblock The jamming transition as a paradigm to understand the loss landscape
  of deep neural networks.
\newblock {\em arXiv preprint arXiv:1809.09349}, 2018.

\bibitem{Franz15}
Silvio Franz, Giorgio Parisi, Pierfrancesco Urbani, and Francesco Zamponi.
\newblock Universal spectrum of normal modes in low-temperature glasses.
\newblock {\em Proceedings of the National Academy of Sciences},
  112(47):14539--14544, 2015.

\bibitem{Wyart05b}
M.~Wyart.
\newblock On the rigidity of amorphous solids.
\newblock {\em Annales de Phys}, 30(3):1--113, 2005.

\bibitem{Liu10}
Andrea J~Liu, Sidney R~Nagel, W~Saarloos, and Matthieu Wyart.
\newblock {\em The jamming scenario - an introduction and outlook}.
\newblock OUP Oxford, 06 2010.

\bibitem{caruana2001overfitting}
Rich Caruana, Steve Lawrence, and C~Lee Giles.
\newblock Overfitting in neural nets: Backpropagation, conjugate gradient, and
  early stopping.
\newblock In {\em Advances in neural information processing systems}, pages
  402--408, 2001.

\bibitem{prechelt1998early}
Lutz Prechelt.
\newblock Early stopping-but when?
\newblock In {\em Neural Networks: Tricks of the trade}, pages 55--69.
  Springer, 1998.

\bibitem{srivastava2014dropout}
Nitish Srivastava, Geoffrey Hinton, Alex Krizhevsky, Ilya Sutskever, and Ruslan
  Salakhutdinov.
\newblock Dropout: a simple way to prevent neural networks from overfitting.
\newblock {\em The Journal of Machine Learning Research}, 15(1):1929--1958,
  2014.

\bibitem{krogh1992simple}
Anders Krogh and John~A Hertz.
\newblock A simple weight decay can improve generalization.
\newblock In {\em Advances in neural information processing systems}, pages
  950--957, 1992.

\bibitem{neyshabur2017geometry}
Behnam Neyshabur, Ryota Tomioka, Ruslan Salakhutdinov, and Nathan Srebro.
\newblock Geometry of optimization and implicit regularization in deep
  learning.
\newblock {\em arXiv preprint arXiv:1705.03071}, 2017.

\bibitem{neyshabur2018towards}
Behnam Neyshabur, Zhiyuan Li, Srinadh Bhojanapalli, Yann LeCun, and Nathan
  Srebro.
\newblock Towards understanding the role of over-parametrization in
  generalization of neural networks.
\newblock {\em arXiv preprint arXiv:1805.12076}, 2018.

\bibitem{bansal2018minnorm}
Yamini Bansal, Madhu Advani, David~D Cox, and Andrew~M Saxe.
\newblock Minnorm training: an algorithm for training over-parameterized deep
  neural networks.
\newblock {\em CoRR}, 2018.

\bibitem{advani2017high}
Madhu~S Advani and Andrew~M Saxe.
\newblock High-dimensional dynamics of generalization error in neural networks.
\newblock {\em arXiv preprint arXiv:1710.03667}, 2017.

\bibitem{belkin2018reconciling}
Mikhail Belkin, Daniel Hsu, Siyuan Ma, and Soumik Mandal.
\newblock Reconciling modern machine learning and the bias-variance trade-off.
\newblock {\em arXiv preprint arXiv:1812.11118}, 2018.

\bibitem{hastie2019surprises}
Trevor Hastie, Andrea Montanari, Saharon Rosset, and Ryan~J Tibshirani.
\newblock Surprises in high-dimensional ridgeless least squares interpolation.
\newblock {\em arXiv preprint arXiv:1903.08560}, 2019.

\bibitem{belkin2019two}
Mikhail Belkin, Daniel Hsu, and Ji~Xu.
\newblock Two models of double descent for weak features.
\newblock {\em arXiv preprint arXiv:1903.07571}, 2019.

\bibitem{liao2018dynamics}
Zhenyu Liao and Romain Couillet.
\newblock The dynamics of learning: A random matrix approach.
\newblock {\em arXiv preprint arXiv:1805.11917}, 2018.

\bibitem{geiger2019scaling}
Mario Geiger, Arthur Jacot, Stefano Spigler, Franck Gabriel, Levent Sagun,
  St{\'e}phane d'Ascoli, Giulio Biroli, Cl{\'e}ment Hongler, and Matthieu
  Wyart.
\newblock Scaling description of generalization with number of parameters in
  deep learning.
\newblock {\em arXiv preprint arXiv:1901.01608}, 2019.

\bibitem{schoenholz2016deep}
Samuel~S Schoenholz, Justin Gilmer, Surya Ganguli, and Jascha Sohl-Dickstein.
\newblock Deep information propagation.
\newblock {\em arXiv preprint arXiv:1611.01232}, 2016.

\bibitem{Tkachenko99}
Alexei~V. Tkachenko and Thomas~A. Witten.
\newblock Stress propagation through frictionless granular material.
\newblock {\em Phys. Rev. E}, 60(1):687--696, Jul 1999.

\bibitem{During13}
Gustavo {D{\"u}ring}, Edan Lerner, and Matthieu Wyart.
\newblock Phonon gap and localization lengths in floppy materials.
\newblock {\em Soft Matter}, 9(1):146--154, 2013.

\bibitem{Wyart05a}
Matthieu Wyart, Leonardo~E Silbert, Sidney~R Nagel, and Thomas~A Witten.
\newblock Effects of compression on the vibrational modes of marginally jammed
  solids.
\newblock {\em Physical Review E}, 72(5):051306, 2005.

\bibitem{saad1995line}
David Saad and Sara~A Solla.
\newblock On-line learning in soft committee machines.
\newblock {\em Physical Review E}, 52(4):4225, 1995.

\bibitem{engel2001statistical}
Andreas Engel and Christian Van~den Broeck.
\newblock {\em Statistical mechanics of learning}.
\newblock Cambridge University Press, 2001.

\bibitem{franzpre}
P.~Urbani S.~Franz, S.~Hwang.
\newblock Jamming in multilayer supervised learning models.
\newblock {\em arXiv preprint arXiv:1809.09945}, 2018.

\bibitem{Franz16}
Silvio Franz and Giorgio Parisi.
\newblock The simplest model of jamming.
\newblock {\em Journal of Physics A: Mathematical and Theoretical},
  49(14):145001, 2016.

\bibitem{Saxe13}
Andrew~M Saxe, James~L McClelland, and Surya Ganguli.
\newblock Exact solutions to the nonlinear dynamics of learning in deep linear
  neural networks.
\newblock {\em International Conference on Learning Representations}, 2014.

\bibitem{Kingma14}
Diederik~P Kingma and Jimmy Ba.
\newblock Adam: A method for stochastic optimization.
\newblock {\em International Conference on Learning Representations}, 2015.

\end{thebibliography}
\bibliographystyle{unsrt}

\appendix

\section{Network properties}
\label{net}
In the following, we analyze numerically the networks properties that were used in the previous analysis. This provides a numerical confirmation of our arguments, and an in depth characterization of the networks.

\subsection{Effective number of degrees of freedom}
\label{app:neff}

Due to several effects discussed above, the function $f(\mathbf{x}; \mathbf{W})$ can effectively depend on less variables that the number of parameters, and thus reduce the dimension of the space spanned by the gradients  $\nabla_\mathbf{W}f(\mathbf{x}; \mathbf{W})$ that enters in the theory. For instance, there could be symmetries that reduce the number of effective degrees of freedom (e.g.\ each ReLU activation function has one of such symmetries, since one can rescale inputs and outputs in such a way that the post-activation is left invariant); another reason could be that a neuron might never activate for all the training data, thus effectively reducing the number of neurons in the network; furthermore, we expect that the network's true dimension would also be reduced if its architecture presents some bottlenecks, is poorly designed or poorly initialized. For example if all biases are too negative on the neurons of one layer in the Relu case, the network does not transmit any signals, leading to $N=1$ and to the possible absence of unstable directions even if the number of parameters is very large. 

It is tempting to define the effective dimension by considering the dimension of the space spanned by $\nabla_\mathbf{W}f(\mathbf{x_\mu}; \mathbf{W})$ as $\mu$ varies. This definition is not  practical for small number of samples $P$ however, because this dimension would be bounded by $P$. We can overcome such a problem by considering a neighborhood of each point $\mathbf{x}_\mu$, where the network's function and its gradient can be expanded in the pattern space:
\begin{equation}
    f(\mathbf{x}) \approx f(\mathbf{x}_\mu) + (\mathbf{x} - \mathbf{x}_\mu) \cdot \nabla_{\mathbf{x}} f(\mathbf{x}_\mu),
\end{equation}
\begin{equation}
    \nabla_\mathbf{W} f(\mathbf{x}) \approx \nabla_\mathbf{W} f(\mathbf{x}_\mu) + (\mathbf{x} - \mathbf{x}_\mu) \cdot \nabla_{\mathbf{x}} \nabla_\mathbf{W} f(\mathbf{x}_\mu).
\end{equation}
Varying the pattern $\mu$ and the point $\mathbf{x}$ in the neighborhood of $\mathbf{x}_\mu$, we can build a family $M$ of vectors:
\begin{equation}
    M = \left\{\nabla_\mathbf{W} f(\mathbf{x}_\mu) + (\mathbf{x} - \mathbf{x}_\mu) \cdot \nabla_{\mathbf{x}} \nabla_\mathbf{W} f(\mathbf{x}_\mu)\right\}_{\mu,\mathbf{x}}.
\end{equation}
We then define the effective dimension $N_\mathrm{eff}$ as the dimension of  $M$. Because of the linear structure of $M$, it is sufficient to consider, for each $\mu$, only $d+1$ values for $x$, e.g.\ $x-x_\mu=0,\hat{\mathbf{e}}_1,\dots,\hat{\mathbf{e}}_d$, where $\hat{\mathbf{e}}_n$ is the unit vector along the direction $n$. The effective dimension is therefore
\begin{equation}
    N_\mathrm{eff} = \mathrm{rk}(G),
\end{equation}
where the elements of the matrix $G$ are defined as
\begin{equation}
    G_{i,\alpha} \equiv \partial_{W_i} f(\mathbf{x}_\mu) + \hat{\mathbf{e}}_n \cdot \nabla_{\hat{\mathbf{e}}_n} \partial_{W_i} f(\mathbf{x}_\mu),
\end{equation}
with $\alpha\equiv(\mu,n)$. The index $n$ ranges from $0$ to $d$, and $\hat{\mathbf{e}}_0 \equiv 0$.

\begin{figure}[htb]
    \centering
    \scalebox{0.7}{\import{figures/}{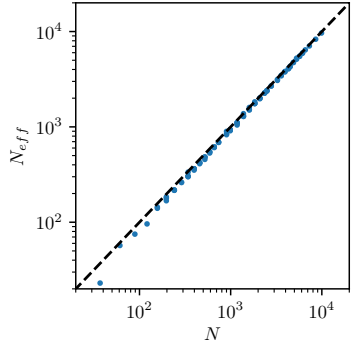}}
    \caption{\small Results with the MNIST dataset, keeping the first 10 PCA components. $d=10$ and $L=3$, varying $P$ and $h$. Effective $N_\mathrm{eff}$ vs total number of parameters $N$. $N_\mathrm{eff}$ is always smaller than $N$ because there is a symmetry per each ReLU-neuron in the network.}
    \label{fig:neff}
\end{figure}

In Fig.~\ref{fig:neff} we show the effective number of parameters $N_\mathrm{eff}$ versus the total number of parameters $N$, in the case of a network with $L=3$ layers trained on the first 10 PCA components of the MNIST dataset. There is no noticeable difference between the two quantities: the only reduction is due to the symmetries induced by the ReLU functions (there is one such symmetry per neuron. Indeed the ReLU function $\rho(z) = z \Theta(z)$ satisfies $\Lambda \rho(z/\Lambda) \equiv \rho(z)$.) We observed the same results for random data.

\subsection{\texorpdfstring{$\mathrm{sp}(H_p)$}{sp(Hp)} is symmetric for ReLu activation functions and random data}
\label{app:hpsym}

We consider $\mathcal{H}_p = -\sum_\mu y_\mu \rho\,(\Delta_\mu)\, \hat{\mathcal{H}}_\mu$, where $\hat{\mathcal{H}}_\mu$ is the Hessian of the network function $f(\mathbf{x}_\mu;\mathbf{W})$ and $\rho$ is the Relu function. We want to argue that the spectrum of $\mathcal{H}_p$ is symmetric in the limit of large $N$. 

We do two main hypothesis: First, the trace of any finite power of $\mathcal{H}_p$ is self-averaging (concentrates) with respect to the average over the random data:
\begin{equation}
\frac{1}{N}\mathrm{tr}(\hat{\mathcal{H}_p}^n) =\frac{1}{N}\overline{\mathrm{tr}(\hat{\mathcal{H}_p}^n) }.
\end{equation}
Second, 
\begin{equation}
\eqalign{\frac{1}{N}\sum_{\mu_1,\cdots,\mu_n} 
\overline{y_{\mu_1}\rho(\Delta_{\mu_1})\cdots y_{\mu_n}\rho(\Delta_{\mu_n})
\mathrm{tr}(\hat{\mathcal{H}}_{\mu_1}\cdots \hat{\mathcal{H}}_{\mu_n}) }
=\\= \frac{1}{N}\sum_{\mu_1,\cdots,\mu_n} 
\overline{y_{\mu_1}\rho(\Delta_{\mu_1})\cdots y_{\mu_n}\rho(\Delta_{\mu_n})} \cdot \overline{
\mathrm{tr}(\hat{\mathcal{H}}_{\mu_1}\cdots \hat{\mathcal{H}}_{\mu_n}) }}
\end{equation}

The first hypothesis is natural since $\hat{\mathcal{H}_p}$ is a very large random matrix, for which the density of eigenvalues is expected to become a non-fluctuating quantity. The second hypothesis is more tricky: it is natural to assume 
that the trace concentrates, however one also need to 
show that the sub-leading corrections to the self-averaging of the trace 
can be neglected. 

Using these two hypothesis and the result, showed below, 
that 
\begin{equation}\label{eqtra}
    \overline{
\mathrm{tr}(\hat{\mathcal{H}}_{\mu_1}\cdots \hat{\mathcal{H}}_{\mu_n}) }=0
\end{equation}
for all $n$ odds, one can conclude that all odds traces of $\hat{\mathcal{H}_p}$ are zero. This implies that 
the spectrum of $\hat{\mathcal{H}_p}$ is symmetric, more precisely that the fractions of negative and positive eigenvalues are equal.  

In order to show that the statement (\ref{eqtra}) above holds, let us argue first that $\overline{\mathrm{tr}(\hat{\mathcal{H}}^n_\mu)} = 0$ for any \emph{odd} $n$.
\begin{equation}
    \mathrm{tr}(\hat{\mathcal{H}}^n_\mu) = \sum_{i_1,i_2,\dots,i_n} \hat{\mathcal{H}}^\mu_{i_1,i_2} \hat{\mathcal{H}}^\mu_{i_2,i_3} \cdots \hat{\mathcal{H}}^\mu_{i_n,i_1},
    \label{eq:tracen}
\end{equation}
where the indices $i_1,\dots,i_n$ stand for synapses connecting a pair of neurons (i.e.\ each index is associated with a synaptic weight $W^{(j)}_{\alpha,\beta}$: we are not writing all the explicit indexes for the sake of clarity). The term of the hessian obtained when differentiating with respect to weights $W^{(j)}_{\alpha,\beta}$ and $W^{(k)}_{\gamma,\delta}$ reads
\begin{equation}
    \eqalign{\hat{\mathcal{H}}^{\mu;(jk)}_{\alpha\beta;\gamma\delta} = \sum_{\pi_0,\dots,\pi_{L}} \theta(a^\mu_{L,\pi_{L}})\cdots\theta(a^\mu_{1,\pi_{1}}) x^\mu_{\pi_{0}} \cdot\\\quad\cdot \partial_{W^{(j)}_{\alpha,\beta}} \partial_{W^{(k)}_{\gamma,\delta}} \left[W^{(L+1)}_{\pi_{L}} W^{(L)}_{\pi_{L},\pi_{L-1}} \cdots W^{(1)}_{\pi_{1}\pi_{0}}\right].}
    \label{eq:hessianfull}
\end{equation}
where we denoted with $a$ the inputs in the nodes of the network. 
Our argument is based on a symmetry of the problem with random data: changing the sign of the weight of the last layer $W^{(L+1)} \longrightarrow -W^{(L+1)}$ and changing the labels $y_\mu\longrightarrow -y_\mu$ leaves the loss unchanged. 
We will show that this symmetry implies that $\mathrm{tr}(\hat{\mathcal{H}}^n_\mu)$ averaged over the random labels is zero for odd $n$. 

In fact, note that the sum in Equation~(\ref{eq:hessianfull}) contains a weight per each layer in the network, with the exception of the two layers $j,k$ with respect to which we are deriving. This implies that any element of the hessian matrix where we have not differentiated with respect to the last layer ($j,k < L+1$) is an odd function of the last layer $W^{(L+1)}$, meaning that if $W^{(L+1)} \longrightarrow -W^{(L+1)}$, then the sign of all these Hessian elements is inverted as well.

If in the argument of the sum in Equation~(\ref{eq:tracen}) there is no index belonging to the last layer, then the whole term changes sign under the transformation $W^{(L+1)} \longrightarrow - W^{(L+1)}$. Suppose now that, on the contrary, there are $m$ terms with one index belonging to the last layer (we need not consider the case of two indices both belonging to the last layer because the corresponding term in the Hessian would be $0$, as one can see in Equation~(\ref{eq:hessianfull})). For each index equal to $L+1$ (the last layer), there are exactly two terms: $\hat{\mathcal{H}}^\mu_{j,L+1} \hat{\mathcal{H}}^\mu_{L+1,k}$ (for some indexes $j,k$). Since $j,k$ cannot be $L+1$ too, this implies that the number $m$ of terms with an index belonging to the last layer is always even. Consequently, when the sign of $W^{(L+1)}$ is reversed, the argument of the sum in Equation~(\ref{eq:tracen}) is multiplied by $(-1)^{n-m}$ (once for each term \emph{without} an index belonging to the last layer), which is equal to $-1$ if $n$ is odd.
The same symmetry can be used to show that a matrix made of an odd product of matrices $\hat{\mathcal{H}}_\mu$, such as $\hat{\mathcal{H}}_\mu \hat{\mathcal{H}}_{\mu'}\hat{\mathcal{H}}_{\mu''}$, must also have a symmetric spectrum,
concluding our argument. 

\begin{figure*}[ht]
    \centering
    \scalebox{0.6}{\import{figures/}{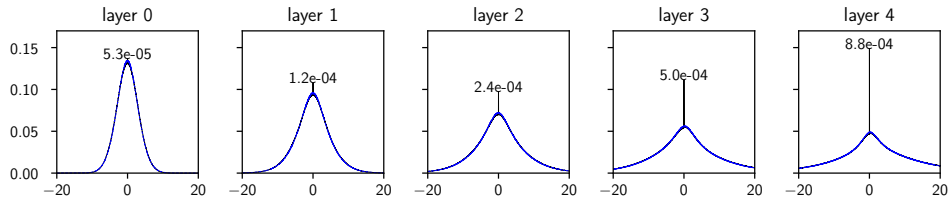}}
    \caption{\small Density of the pre-activations for each layers with $L=5$ and random data, averaged over all the runs just above the jamming transition with that architecture. Black:  distribution obtained over the training set. Blue: previously unseen random data (the two curves are on top of each other except for the delta in zero). The values indicate the mass of the peak in zero, which is only present when the training set is considered.}
    \label{fig:preactivity}
\end{figure*}

\subsection{Density of pre-activations for ReLU activation functions}
\label{app:zeros}

The densities of pre-activation (i.e. the value of the neurons before applying the activation function) is shown in Fig.~\ref{fig:preactivity} for random data. It contains a delta distribution in zero. The number $N_c$ of pre-activations equal to zero when feeding a network $L=5$ all its random dataset is $N_c\approx 0.21 N$, corresponding to the number of directions in phase space where cusps are present in the loss function. For MNIST data we find $N_c\approx 0.19N$. By taking $L=2$ and random data we find $N_c\approx 0.25N$. In these directions, stability can be achieved even if the hessian would indicate an instability. For this reason, instead of $N_-$ in Equation~(\ref{4bis}) one should use $N/2-N_c\approx0.25 N$.

\section{Parameters used in simulations}
\label{app:simulations}

\subsection{Random data}

The dataset is composed of $P$ points taken to lie on the $d$-dimensional hyper-sphere of radius $\sqrt{d}$, ${\bf x}_\mu\in {\cal S}^d$, with random label $y_\mu=\pm 1$. The networks are fully connected, and have an input layer of size $d$ and $L$ layers with $h$ neurons each, culminating in a final layer of size $1$. To find the transition we proceed as follows: we build a network with a number of parameters $N$ large enough for it to be able to fit the whole dataset without errors. Next, we decrease the width $h$ while keeping the depth $L$ fixed, until the network cannot correctly classify all the data anymore within the chosen learning time. We denote this transition point $N^*$.
As initial conditions for the dynamics we use the default initialization of {\tt pytorch}: weights and biases are initialized with a uniform distribution on $[-\sigma, \sigma]$, where $\sigma^2 = 1/f_{in}$ and $f_{in}$ is the number of incoming connections.

\begin{figure}[ht]
    \centering
    \scalebox{0.6}{\import{figures/}{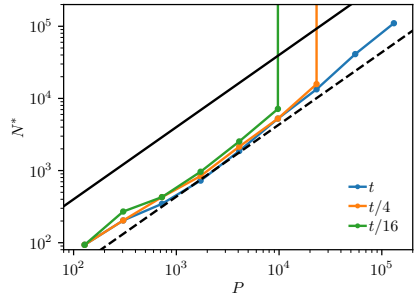}}
    \caption{\small Convergence of the critical line for networks trained with cross entropy on random data.}
    \label{fig:xent_convergence}
\end{figure}

When using the cross entropy, the system evolves according to a stochastic gradient descent (SGD) with a learning rate of $10^{-2}$ for $5\cdot10^5$ steps and $10^{-3}$ for $5\cdot10^5$ steps ($10^6$ steps in total); the batch size is set to $\min(P/2, 1024)$, and batch normalization is used. We do not use any explicit regularization in training the networks. In Fig.~\ref{fig:xent_convergence} we check that $t=10^6$ is enough to converge.

When using the hinge loss, we use an orthogonal initialization~\cite{Saxe13}, no batch normalization and $t=2\cdot10^6$ steps of ADAM~\cite{Kingma14} with batch size $P$ and a learning rate starting at $10^{-4}$. In the experiments of section \ref{sec:random} (not for the experiments of section~\ref{sec:gen}), we progressively divided the learning rate by $10$ every 250k steps. Also in this case we do not use any explicit regularization in training the networks.

To observe the discontinuous jump in the number $N_\Delta$ of unsatisfied constraints at the transition (Fig.~\ref{fig:fit}B and inset), we consider three architectures, both with $N \approx 8000$ and $d=h$ but with different depths $L=2$, $L=3$ and $L=5$. The vicinity of the transition is studied by varying $P$ around the transition value and minimizing for $10^7$ steps (a better minimization is needed to improve the precision close to the transition).

\paragraph{Details about Fig~\ref{fig:fit}A hinge} We took $d=h$ and trained for 2M steps. For some values of $P \in (500, 60\mathrm{k})$, start at large $h$ where we reach $N_\Delta = 0$ and decrease $h$ until $N_\Delta > 0.1N$.

\paragraph{Details about Fig~\ref{fig:fit}B}
We trained networks of depth 2,3,5 with $d=h=$ 62, 51, 40 respectively for 10M steps.
For $L=3$ ($d=51$, $h=51$) we ran 128 training varying $P$ from 21991 to 25918. 
For the value of $N$ we take $7854$ that correspond to the number of parameters minus the number of neurons, per neuron there is a degree of freedom lost in a symmetry induced by the homogeneity of the ReLU function.
37 of the runs have $N_\Delta = 0$, 74 have $N_\Delta > 0.4N$. Among the 19 remaining ones, 14 of them have $N_\Delta$ between 1 and 4, we think that these runs encounter numerical precision issues, we observed that using 32 bit precision accentuate this issue.
We think that the 5 left with $4 < N_\Delta < 0.4 N$ has been stoped too early. The same observation apply for the other depths.


\subsection{Real data}\label{app:realdatadepths}

The images in the MNIST dataset are gathered into two groups, with even and odd numbers and with labels $y_\mu=\pm1$. The architecture of the network is as in the previous sections: the $d$ inputs are fed to a cascade of $L$ fully-connected layers with $h$ neurons each, that in the end result in a single scalar output. The loss function used is always the hinge loss.

If we kept the original input size of $28\times28=784$ (each picture is $28\times28$ pixels) then the majority of the network's weights would be necessarily concentrated in the first layer (the width $h$ cannot be too large in order to be able to compute the Hessian). To avoid this issue, we opt for a reduction of the input size. We perform a principal component analysis (PCA) on the whole dataset and we identify the 10 dimensions that carry the most variance on the whole dataset; then we use the components of each image along these directions as a new input of dimension $d=10$. This projection hardly diminishes the performance of the network (which we find to be larger than $90\%$ when using all the data and large $N$).

\paragraph{Details about Fig~\ref{fig:fit}C}
We trained networks of depth 1,3,5 for 2M steps.
For some values of $P \in (100, 50\mathrm{k})$, start at large $h$ where we reach $N_\Delta = 0$ and decrease $h$ until $N_\Delta > 0.1N$.

\paragraph{Details about Fig~\ref{fig:fit}D}
We trained a network of $L=5$, $d=10$, $h=30$ for 3M steps.
With $P$ varying from 31k to 68k (using trainset and testset of MNIST).

\begin{figure}[htb]
    \centering
    \scalebox{0.7}{\import{figures/}{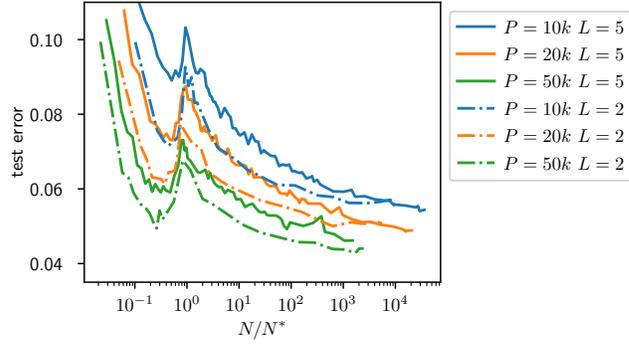}}
    \caption{\small Generalization on MNIST 10 PCA. Comparison between two depth $L=2$ and $L=5$.}
    \label{fig:gen_depth}
\end{figure}

\paragraph{Details about Fig~\ref{fig:gen}}
We trained a network of $L=5$ and $d=10$ for 500k steps.
where $P \in \{10\mathrm{k}, 20\mathrm{k}, 50\mathrm{k}\}$ and $h$ varies from 1 to 3k.
Fig~\ref{fig:gen_depth} shows a comparison between $L=5$ and $L=2$.

\end{document}